\documentclass[12pt]{article}
\usepackage{fullpage,graphicx,psfrag,amsmath,amsfonts,verbatim}

\usepackage[small,bf]{caption}
\usepackage{url,amsthm}
\usepackage{subcaption}
\usepackage{lscape} 
\usepackage{booktabs}
\usepackage{placeins}
\usepackage{fancyvrb}

\newcommand{\BEAS}{\begin{eqnarray*}}
\newcommand{\EEAS}{\end{eqnarray*}}
\newcommand{\BEQ}{\begin{equation}}
\newcommand{\EEQ}{\end{equation}}
\newcommand{\BIT}{\begin{itemize}}
\newcommand{\EIT}{\end{itemize}}

\newcommand{\eg}{{\it e.g.}}
\newcommand{\ie}{{\it i.e.}}

\newcommand{\reals}{{\mbox{\bf R}}}
\newcommand{\integers}{{\mbox{\bf Z}}}
\newcommand{\naturals}{{\mbox{\bf N}}}
\newcommand{\symm}{{\mbox{\bf S}}}

\newcommand{\argmax}{\mathop{\rm argmax}}

\newcommand{\na}{\texttt{nan}}

\usepackage{hyperref} 
\usepackage{amsthm}

\usepackage[toc,page]{appendix}
\begin{document}

\title{Seasonally-Adjusted Auto-Regression of
Vector Time Series}
\author{
Enzo Busseti \\
\small {Department of Management Science and Engineering}\\
\small {Stanford University}
}

\date{\today}

\maketitle
\begin{abstract}
We present a simple algorithm
to forecast vector time series, that is
robust against missing data, in both 
training and inference.
It models seasonal  
annual, weekly, and daily baselines,
and a Gaussian process for the seasonally-adjusted
residuals.
We develop a custom truncated eigendecomposition
to fit a low-rank plus block-diagonal Gaussian kernel. 
Inference is performed with the Schur complement,
using Tikhonov regularization to prevent overfit,
and the Woodbury formula
to invert sub-matrices of the kernel efficiently.
Inference requires an amount of memory and computation
linear in the dimension of the time series,
and so the model can
scale to very large datasets.
We also propose a simple ``greedy'' grid search 
for automatic hyper-parameter tuning.
The paper is accompanied by \verb+tsar+
(\ie, time series auto-regressor),
a Python library that implements the algorithm.
\end{abstract}


\section{Introduction}

We present a model, accompanied by a software
implementation, for a time series. This
can be used to forecast future unknown values of the series.

\paragraph{Note for the reader.}
The present paper is a draft that still lacks 
references,
examples, 
and applications of the model.

\paragraph{Notation.}
We use some simple notational conventions.
Whenever a variable has a subscript ``$t$''
it is a time series and $t$ indexes time,
for example, $b_t \in \reals^M$ for
$t \in \integers$, is a (real) vector time series.
Each value of $t$ corresponds to a point in time,
and the time interval between $t$ and $t+1$ is
the same for each value of $t$.
For example, in a \emph{hourly} time series
each value of $t$ typically corresponds to the start of each hour.
The choice of the origin for the time index is arbitrary.
Whenever we write a variable with a ``hat'' on top
we mean it is a statistical inference of the variable
without it.
In addition, if we are inferring a variable in a time series, 
by the notation $\hat x_{\tau | t}$ we mean
the inferred value of variable $x_\tau$ 
using the data available at time $t$.
If $t < \tau$ then $\hat x_{\tau | t}$ is a
\emph{forecast} or prediction,
if $t \geq \tau$ then $\hat x_{\tau | t}$ is called
a \emph{nowcast} or \emph{imputation} (of missing values).
Finally, we denote by $\na$ a missing value 
 of a scalar variable
 (literally, ``not a number'').
Typically, a scalar variable in a raw dataset takes value in 
$\reals \cup \{\na\}$,
meaning that it is either a real 
number or is missing.
We use the simple algebraic convention that 
$a + \na = a - \na = a \times \na = a / \na = \na$
for any $a \in \reals \cup \{\na\}$. 
We treat the problems of forecasting future values 
and of guessing or nowcasting missing values as the same. 

\paragraph{Our objective.}
We consider a vector time series
$x_t \in {(\reals \cup \{\na\})}^M$ 
for $t \in \integers$ and $M \geq 1$.
We use the notation
$x_{t,i} \in \reals \cup \{\na\}$, 
for any $i = 1, \ldots M$,
to mean the $i$-th element
of the $t$-th observation of the series.
We are given data for a certain period, 
\[
x_{t_s}, \ldots, x_{t_e},
\]
 where $t_e > t_s$
are its end and start times. 
The provided data can have any number
of missing values,
and
the time series is equal to $M-$vectors of missing values
outside of the interval
\[
x_\tau = (\na, \ldots, \na)
\text{ if } 
\tau < t_s
\text{ or } 
\tau > t_e.
\]
Our objective is
to fill the missing values of the time series.
The machinery we build can be used
to fill any missing value:
we can guess the time series at times
before $t_s$,
impute the missing entries in the data provided,
and forecast the future values after $t_e$.
In the rest of the paper, 
for ease of explanation,
we focus on the case of 
producing forecasts of the future values
\[
\hat x_{\tau | t_e} 
\in \reals^M,
\quad
\tau > t_e.
\]

\paragraph{The model.}
We model the time series as
\[
x_t = b_t + r_t,
\]
 the sum of a seasonal \emph{baseline} $b_t \in \reals^M$ 
and 
an auto-regressive \emph{residual}
$r_t \in {(\reals \cup \{\na\})}^M$,
for all $t \in \integers$.
The baseline has always real values, 
the residual instead has a missing value
wherever the time series has one.
The baseline explains periodic patterns with
daily, weekly, and annual seasonalities, and 
an optional linear trend. 
(Other periodicities
such as quarterly 
or lunar
could be also included,
but in this work we only focus on those three.)
The baseline separates across the components of the time
series, \ie, it is composed of $M$ separate scalar baseline
functions. We show our proposed baseline function in \S\ref{sec:baseline}.
The residual represents instead a mean-reverting
deviation from the baseline. 
We propose to model it as 
a partially observed zero-meaned Gaussian process with
a finite memory. 
The missing values are simply the unobserved values 
of the process, and
we fill them by computing
their Gaussian conditional expectations,
given the observed values.
We propose various computational optimizations
to achieve better performance in practical usage,
such as a low-rank approximation of the Gaussian kernel
that ensures that the computational cost 
of inferring the missing values is linear,
as opposed to cubic, in $M$.  
We detail the Gaussian process model in 
\S\ref{sec:residuals}.
Both components of the model contain various 
\emph{hyper-parameters}
that an expert user may control
to achieve better performance. 
We propose a simple heuristic
to choose automatically good values for these, 
a \emph{greedy} grid search, 
where ``greedy'' has the meaning
given to it in computer science:
it iteratively tries
small changes in the hyper-parameters,
myopically selecting the ones giving 
better improvements.
We explain it in \S\ref{sec:ggs}.   



\paragraph{Train-test split.}
Throughout the paper we detail procedures
to fit models using \emph{train} data
$x_{t^\text{tr}_s}, \ldots, x_{t^\text{tr}_e}$,
and to evaluate models using \emph{test} data
$x_{t^\text{te}_s}, \ldots, x_{t^\text{te}_e}$. 
These are two subsets of the initial dataset
$x_{t_s}, \ldots, x_{t_e}$,
split by a simple rule:
With a user-defined ratio $r \in (0,1)$, 
by default $r = 2/3$, we choose 
$t^\text{tr}_s = t_s$,
$t^\text{te}_s = t^\text{tr}_e + 1$,
$t^\text{te}_e = t_e$,
and $(t^\text{tr}_e - t^\text{tr}_s) \approx r (t^\text{te}_e - t^\text{te}_s )$,
where the last equation is valid to the closest approximation
possible.
That is, we use approximately
the first $r$ fraction of the data as train data,
and the rest as test data.
For simplicity, the train and test datasets have
 values all equal to $\na$ outside of their boundaries.
The test data is used to select the values of the hyper-parameters.
Once these have been chosen, each model is re-fit 
on the complete original dataset.

\paragraph{Organization of the paper.}
We divide the paper in sections that correspond to sub-modules
of the \verb+tsar+ library , so that each section can
in principle be thought of as a separate document. 
However, the order in which we present them matters, 
since each section depends on concepts from,
or is motivated by, the preceding ones.
The last section is an exception (as is this introduction):
it describes the software implementation
 and exemplifies its use.

\paragraph{Originality.}
None of the material presented in this paper is completely new.
For example, 
the predictive model explained in 
\cite[Appendix A]{MBBW19}
is a special case of the present model, 
for a scalar time series with no missing values,
and no logic to automatically select the hyper-parameter
values.
The idea of separating a time series in a seasonal
and an auto-regressive component is very
old,
it is sometimes called seasonal auto-regression (SAR).
Projecting a periodic function on sines and cosines
up to a certain frequency is a very common
basis expansion, and is at the core of Fourier analysis.
Using Gaussian processes to generalize  
auto-regressive models is also not a new idea,
and low-rank approximations,
such as principal component analysis (PCA),
are widely used in practice.
Tikhonov regularization, which we use to control
overfit of the Gaussian kernel, is also a classic idea.
Lastly, the grid search of hyper-parameters
is a typical procedure
and greedy searches are explained in any
introductory computer science course.
To our knowledge, however, 
no published material
models time series as sums of
truncated Fourier expansions and Gaussian processes, 
the low-rank plus block-diagonal Gaussian kernel
we propose has not appeared before,
nor has the greedy grid search idea.
In addition, we wrote all the software implementation,
with a focus on simplicity and usability.

\section{Seasonal baseline}
\label{sec:baseline}
Here we develop a model for the baseline
$b_t \in \reals^M$ for any $t \in \integers$,
that separates on the $M$ components
of the time series. So, for simplicity of notation, 
and without loss of generality,
in this section we fix $M = 1$, so
$x_t \in \reals \cup \{ \na \}$
and $b_t \in \reals$ for any $t \in \integers$.

We propose, as in 
\cite{MBBW19},
to represent the baseline as a sum
of sine and cosine basis functions,
so as to
capture variations with 
periodically repeating patterns.
Other strategies can be used, 
as long as they are robust
against missing data. For example,
we include in our software library the
non-parametric baseline of \cite{BB19}
as an option.
We model the baseline 
as the sum
\begin{multline}
\label{eq:baseline}
b_t = 
K^\text{trend} \alpha_0 t + \beta_0 \\
+
\sum_{k = 1}^{K^\text{day}} 
\alpha^\text{day}_k \sin (2 \pi t k / P^\text{day}) + 
\beta^\text{day}_k \cos (2 \pi t k / P^\text{day}) \\
+
\sum_{k = 1}^{K^\text{week}} 
\alpha^\text{week}_k \sin (2 \pi t k / P^\text{week}) + 
\beta^\text{week}_k \cos (2 \pi t k / P^\text{week})\\
+
\sum_{k = 1}^{K^\text{year}} 
\alpha^\text{year}_k \sin (2 \pi t k / P^\text{year}) + 
\beta^\text{year}_k \cos (2 \pi t k / P^\text{year})\\
\end{multline}
where $P^\text{day} \in \reals$, $P^\text{week}\in \reals$, and
$P^\text{year}\in \reals$ are the lenghts of
a day, week, and year, in the time interval spacing of the time 
series (\eg, if the series is hourly,
 $P^\text{day} = 24$,
$P^\text{week} = 168$, and $P^\text{year} = 8766$), 
and 
$K^\text{trend} \in \{0, 1\}$,
$K^\text{day} \in \{0, \ldots\}$,
$K^\text{week} \in \{0, \ldots, 6\}$,
$K^\text{year} \in \{0, \ldots, 51\}$
are hyper-parameters with a range chosen so that
the periodicities of the terms in the baseline are unique,
and $\alpha_0$, $\beta_0$, 
$\alpha_k^\text{day}$, $\beta_k^\text{day}$,
$\alpha_k^\text{week}$, $\beta_k^\text{week}$,
$\alpha_k^\text{year}$, $\beta_k^\text{year}$,
for any $k$,
are the coefficients or parameters of the model,
and they are all real numbers.
In the following we use the shorthand notation 
\[
b_t(\alpha, \beta; K),
\]
where $\alpha$, $\beta$, and $K$, are the vectors
of parameters and hyper-parameters 
for all subscripts and superscripts.
So, we use as periods the fundamental periods of one day, 
one week, one year,
and their first few harmonics, \ie, half of each period, a third of,
and so on.
The numbers of harmonics used, or whether a fundamental
 period is used
at all, are chosen by fixing the values of hyper-parameters,
which thus control the complexity of the model.
In addition, we have a constant term $\beta_0$ and a linear
trend $\alpha_0 t$ which is only active if 
$K^\text{trend} = 1$.
The number of effective coefficients of the model
is $2(K^\text{day} + K^\text{week} + K^\text{year})
 + K^\text{trend} + 1$
 (since $\alpha_0$ is irrelevant when $K^\text{trend} = 0$).

\paragraph{Fit.}
We detail an ad-hoc procedure to fit the baseline model,
\ie, to obtain the values of the coefficients $\alpha$ and $\beta$,
given a sequence $x_{t_s}^\text{tr}, \ldots, x_{t_e}^\text{tr}$ 
of \emph{train} data, 
and chosen values of the hyper-parameters $K$.
We use simple least-squares optimization, minimizing
the total squared deviation between the baseline model
and the given data. 
The coefficients are the solution of the following 
least-squares optimization problem
\BEQ
\label{eq:baselinefit}
\begin{array}{rl}
\text{minimize} & 
\sum_{t \in \{t_s^\text{tr},\ldots, t_e^\text{tr} \mid  x_t \neq \na \}}
\|
x_t - b_t(\alpha, \beta; K)
\|^2_2 
 + \gamma  \left(\|\alpha\|_2^2 + \|\beta\|_2^2 
 - \beta_0^2 
 \right)
  \\
\end{array}
\EEQ
where the optimization variables are $\alpha$ and $\beta$,
for all subscripts and superscripts, 
and $\gamma >0$ is a regularization constant
used to ensure uniqueness of the solution, 
unless
all data is missing in the train dataset,
in which case we also fix $\beta_0 = 0$. 
We subtract $\beta_0$ from the regularization term
to ensure that
the deviation between the baseline and the train data
has mean zero.
We fix the value of $\gamma$ to a small constant,
by default $10^{-8}$ in the software package.

\paragraph{Infer.}
Given values of the hyper-parameters $K$,
and the coefficients $\alpha$ and $\beta$,
to infer the value of the baseline at any time $t \in \integers$
we simply evaluate it
\[
\hat b_t = b_t(\alpha, \beta; K).
\] 
Thus inference can be performed at any time, also, \eg,
 in the distant future. For time series whose long-term
  dynamics is not expected to change much in time, such 
  as the energy production of a renewable power plant which depends
  only on weather, the baseline model by itself
  can be used as a (rough) estimate of the far future.

 \paragraph{Evaluate.}
 It is easy to evaluate the goodness of 
 a baseline model, \ie, its
set of parameters $\alpha$ and $\beta$
 and hyper-parameters $K$,
  over some \emph{test} data
 $x_s^\text{te}, \ldots, x_e^\text{te}$,
 by the squared deviation
  \BEQ
\label{eq:baselineeval}
\sum_{t \in \{t_s^\text{te},\ldots, t_e^\text{te} \mid  x_t \neq \na \}}
\|
x_t - \hat b_t
\|^2_2,
\EEQ
skipping missing data.
The smaller the value of \eqref{eq:baselineeval},
the better. The greedy grid search we use to
choose the values of the hyper-parameters $K$ relies
on empirically minimizing \eqref{eq:baselineeval}, with
the original data $x_s, \ldots, x_e$
split between train and test datasets.




\section{Gaussian process of the residuals}
\label{sec:residuals}
Once a baseline model has been fitted, we subtract it
from the original data to get a time series of residuals
\[
r_t = x_t - b_t
\]
where
$r_t \in {(\reals \cup \{ \na\})}^M$,
for any $t\in\integers$.
In particular, for any $t \in \integers$
and $i =1, \ldots, M$,
if $x_{t,i} = \na$ 
then $r_{t,i} = \na$.

\paragraph{Normalization.}
We divide the residuals by their empirical norm.
Let $\sigma \in \reals^M$ be defined as
\[
 \sigma_i = 
\sqrt{
\frac{
\sum_{t \in 
\{t= t_s^\text{tr}, \ldots, t_e^\text{tr} \mid r_{t,i} \neq \na\}}
r_{t,i}^2}
{
\sum_{t \in 
\{t= t_s^\text{tr}, \ldots, t_e^\text{tr} \mid r_{t,i} \neq \na\}}
1
}}
\]
for $i = 1, \ldots, M$,
and $\sigma_i = 1$ if all
$r_{t_s^\text{tr}, i}, \ldots, r_{t_e^\text{tr}, i}= \na$.
Then the normalized residuals are
\BEQ
\label{eq:normresidual}
\tilde r_{t,i} = r_{t,i} / \sigma_i
\EEQ
for any $t \in \integers$ and $i = 1, \ldots, M$.
We model $\tilde r_t$ as 
a Gaussian process with a kernel size
defined by the user. 
This model
generalizes the classic auto-regression
of a time series, which is
a special case when there are no missing data.

\paragraph{Gaussian kernel.}
Let $P \in \naturals$ be the lenght of
the \emph{past}, or \emph{memory},
of the kernel,
and 
$F \in \naturals$ be the lenght of
the \emph{future}, or forecast \emph{horizon}.
These are specified by the user, 
based on the requirements of the application
of the model. For example, if the time series is comprised
of hourly data, one could choose $P = F = 24$,
or $P = F = 48$.
We model the normalized residual as a partially observed
Gaussian process with mean zero,
so that for any $t \in \integers$
\BEQ
\label{eq:multivariate_gauss}
(\tilde r_{t - P+1,1} , \ldots, \tilde r_{t + F,1},
\ldots
\tilde r_{t - P+1,M} , \ldots, \tilde r_{t + F,M})
\sim
\mathcal{N}(0, \Sigma)
\EEQ
where
$\Sigma \in \symm_{++}^{(P + F)M}$
is the Gaussian \emph{kernel} or covariance,
a real positive definite matrix,
and the $\sim$ operator is overloaded
to mean that the variables on the left
are distributed according to the multivariate 
Gaussian distribution on the right,
but are only partially observed, \ie, 
can have missing values.

\begin{center}
\begin{figure}
\includegraphics[width=\textwidth]{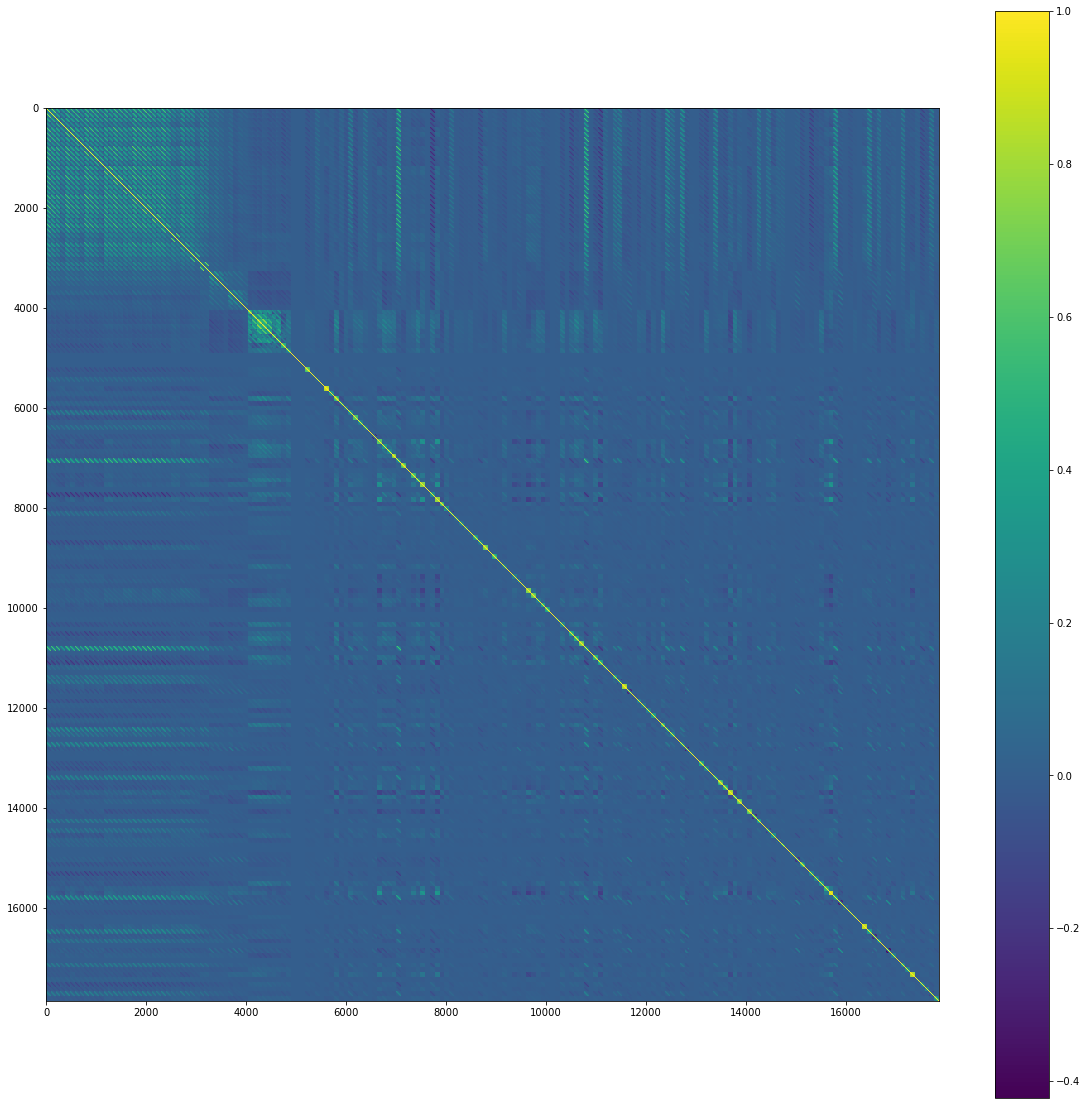}
\caption{The heatmap plot of an example low-rank plus
block diagonal kernel. The data used to estimate it
involve a few hundred measured and forecasted quantities
related to a portfolio management problem. 
The brigher the color, 
the stronger the correlation between a pair of variables.
The diagonal blocks modeling the auto-correlation
of each separate variable are clearly visible.
All estimation and inference procedures with this kernel
take, respectively, a few seconds and about one, 
on a consumer laptop.}
\label{fig:kernel_heatmap}
\end{figure}
\end{center}

\paragraph{Fit.}
We fit an approximate Gaussian kernel
$\hat \Sigma \in \symm^{(P + F)M}$ as follows.
The kernel is divided in submatrices
\BEQ
\label{eq:gaussian_kernel_full}
\hat \Sigma = 
\left(
\begin{array}{ccc}
C_{(1,1)} & \cdots & C_{(1,M)}\\
\vdots & \ddots &\vdots \\
C_{(M,1)} & \cdots & C_{(M,M)}\\
\end{array}
\right).
\EEQ
where, for any $i, j = 1, \ldots, M$,
the matrix 
$C_{(i,j)} \in \reals^{(P+F) \times (P+F)}$
is Toeplitz 
\[
C_{(i,j)} = 
\left(
\begin{array}{cccc}
c^{(i,j)}_0 & c^{(i,j)}_1 & \cdots &  c^{(i,j)}_{P+F - 1} \\
c^{(i,j)}_{-1} &\ddots  &  & \vdots \\
\vdots &&  & c^{(i,j)}_{1}\\
c^{(i,j)}_{1-P-F} & \cdots& c^{(i,j)}_{-1} &c^{(i,j)}_0
\end{array}
\right).
\]
The approximate \emph{correlation} coefficients
$c^{(i,j)}_\tau \in \reals$, for $\tau = 1-P-F, \ldots, P+F-1$,
are defined as
\BEQ
\label{eq:corrcoeff}
c^{(i,j)}_\tau = 
\frac{
\sum_{t \in 
\{t= t_s^\text{tr}, \ldots, t_e^\text{tr} \mid r_{t,i} \neq \na, r_{t+\tau,j} \neq \na\}}
\tilde r_{t,i} \tilde r_{t+\tau, j}}
{
\sum_{t \in 
\{t= t_s^\text{tr}, \ldots, t_e^\text{tr} \mid r_{t,i} \neq \na, r_{t+\tau, j} \neq \na\}}
1
},
\EEQ
or $c^{(i,j)}_\tau = 0$ if the denominator in \eqref{eq:corrcoeff}
is equal to 0,
and we remind that the train dataset has values all equal
to $\na$ when $t < t_s^\text{tr}$ 
and $t > t_e^\text{tr}$.
The approximate kernel so constructed is symmetric
but is not in general positive definite,
unless there are no missing data,
in which case $\hat \Sigma$ is at least
positive semi-definite
(indeed, an empirical correlation matrix).
We use Tikhonov regularization during inference
to correct the non-positive-definiteness of $\hat \Sigma$.
We show an example estimated kernel,
as visualized by a heatmap of the kernel matrix values,
in Figure \ref{fig:kernel_heatmap}.

\paragraph{Infer.}
We infer the missing values of the normalized
residual, and hence of the residual,
by computing their Gaussian conditional expectations.
For a given time index $t \in \integers$
we concatenate the normalized residuals
as in \eqref{eq:multivariate_gauss}
to obtain
\[
\label{eq:concat}
\tilde \rho = (\tilde r_{t -P+1, 1}, \ldots, \tilde r_{t+F, 1},
\ldots,
\tilde r_{t -P+1, M}, \ldots, \tilde r_{t+F, M})
\in {(\reals \cup \{\na\})}^{M(P+F)}.
\]
We form two sets of indexes of 
$\tilde \rho$.
Let
$\mathcal{O}$ be the set 
of indexes of \emph{observed} values,
and 
$\mathcal{U}$ the set 
of indexes of \emph{unobserved} values.
Together they are
a partition of the set
$\{1, \ldots,M(P+F)\}$
of indexes of $\tilde \rho$.
For any $i \in \mathcal{O}$ we have
$\tilde \rho_i \in \reals$,
and for any $i \in \mathcal{U}$,
$\tilde \rho_i = \na$.
Then, given an estimated kernel $\hat \Sigma \in \symm^{M(P+F)}$
and a Tikhonov regularization parameter
$\lambda \geq 0$,
inference is performed via the
regularized Schur complement
\BEQ
\label{eq:schur}
\hat{\tilde{\rho}}_\mathcal{U}
= 
\hat\Sigma_{\mathcal{U},\mathcal{O}}
{(\hat\Sigma_{\mathcal{O},\mathcal{O}} + \lambda I)}^{-1} 
\tilde{\rho}_\mathcal{O},
\EEQ
where $I$ is an identity matrix of appropriate size
and $\lambda$ is chosen large enough so that
the inverse matrix above exists and is real.
For simplicity, we define 
$\hat{\tilde{\rho}}_\mathcal{O} = \tilde{\rho}_\mathcal{O}$,
so the vector $\hat{\tilde{\rho}} \in \reals^{M(P+F)}$ 
is well defined.
We deconcatenate it,
obtaining 
$\hat{\tilde{r}}_{t-P+1}, \ldots,\hat{\tilde{r}}_{t+F}$,
and finally we get the inferred unnormalized
residuals by
inverting \eqref{eq:normresidual}
\[
\hat r_{\tau, i} = \sigma_i  \hat{\tilde{r}}_{\tau, i}
\]
for any $i =1, \ldots, M$ and
$\tau = t-P+1, \ldots, t+F$.
The regularization coefficient
$\lambda$ is thus an hyper-parameter of the model,
and for simplicity we give it a range
$\lambda \in \{
M(P+F),
M(P+F)/\alpha,
M(P+F)/\alpha^2,
\ldots
\}$,
where $\alpha$ is a constant
(by default $\alpha = \sqrt[3]{10}$ in the software).

\paragraph{Auto-regression.}
If the kernel is estimated on data without any missing
value, the inference is exclusively performed
on data such that
$r_{t-P+1}, \ldots, r_t \in \reals^M$,
and   
$r_{t+1}, \ldots, r_{t+F} \in {(\{\na\})}^M$,
and finally $\lambda = 0$,
then the procedure explained above is equivalent to 
a classic vector auto-regression with memory $P$
and prediction horizon $F$,
see for example \cite[Appendix A]{MBBW19}.
(Some authors define the auto-regression only with $F = 1$.)
That is in turn equivalent to a linear regression
of the observations of $\tilde r$
at times $t+1, \ldots, t+F$ on
the observations at times
$t-P+1, \ldots, t$ for every $t$,
with no intercept, because their mean is zero.
If $\lambda > 0$, then the procedure is equivalent to a 
\emph{ridge} regression.
Our approach is more flexible than a classic 
auto-regression since it does not assume
a fixed pattern of observed data, and hence can
handle any past missing, or future present, value.

\paragraph{Evaluate.}
We evaluate the performance of a fitted model
by the
sum of the squared deviations between the model predictions 
of the future, and the real values, on a test dataset.
First, given the test dataset
$
r_{t_s^\text{te}}, \ldots, r_{t_e^\text{te}}
$,
we normalize by the $\sigma$ computed on the train
dataset to obtain the normalized residuals
$\tilde r_{t_s^\text{te}}, \ldots, \tilde r_{t_e^\text{te}}$.
Then, for every $t= t_s^\text{te}, \ldots, t_e^\text{te}$,
and 
$\tau = t + 1, \ldots, t+F$,
we use the notation
$\hat{\tilde{r}}_{\tau | t}$,
to mean the inferred value of  
$\tilde{r}_{\tau}$
computed according to
\eqref{eq:schur}
with all normalized residuals at times 
after $t$ set equal to $\na$,
\ie,
$\tilde r_{t+1} = \ldots = \tilde r_{t+F}=  (\na, \ldots, \na)$.
Finally, we measure the performance by
\[
\sum_{t = t_s^\text{te}}^{t_e^\text{te}}
\sum_{i = 1}^M
\sum_{\tau \in \{t+1, \ldots, t+F
 \mid r_{\tau,i}\neq \na \}}
(\hat{\tilde{r}}_{\tau | t,i} - \tilde r_{\tau,i})^2.
\]

\paragraph{Memory and computational cost.}
The memory cost of storing $\hat \Sigma$
is $O(M^2(P+F))$,
quadratic in $M$ and linear in $P+F$ (because
each sub-matrix is Toeplitz).
The computational cost of fitting
 $\hat \Sigma$ is approximately
 $O(M^2(P+F)T)$, where $T$
 is the number of observations in the dataset,
 so quadratic in $M$, linear in $P+F$,
 and linear in the size of the dataset.
 The cost of inference, in the worst case, is
 $O(M^3(P+F)^3)$,
    the matrix inverse computation, so it is
 cubic in $M$ and $(P+F)$.
 We can however cache the matrix 
 $\hat\Sigma_{\mathcal{U},\mathcal{O}}
{(\hat\Sigma_{\mathcal{O},\mathcal{O}} + \lambda I)}^{-1}$
and re-use it if we need to predict with
the same pattern of observed and unobserved data.
In that case, the cost of re-use would
 only be quadratic in $M$ and $P+F$,
 as would be the cost of storing the matrix.



\subsection{Low-rank plus block diagonal kernel}
We now develop an approximation of the 
estimated Gaussian kernel $\hat \Sigma$
defined in \eqref{eq:gaussian_kernel_full}
that gives a great computational speedup,
so that the computation cost of 
inference is linear, rather than cubic, in $M$.
The main idea is to obtain a small number $R \geq 0$
of \emph{principal directions} 
$v_1,\ldots, v_R  \in \reals^M$,
with $R < M$,
that explain
most of the changes in $\tilde r_t$, model
the auto-regression along such directions,
and neglect the other joint variations of the residual series.
We  retain however
the (scalar) auto-regressive model of each of the $M$  components
of $\tilde r_t$.
So, $R$ is the second hyper-parameter
of the Gaussian process, and has range
$R \in \{0, \ldots, M\}$.

\paragraph{Principal directions.}
We consider the eigendecomposition
of the matrix of approximate correlation
coefficients of the variables in the normalized residual,
 defined in \eqref{eq:corrcoeff}
\[
\left(
\begin{array}{ccc}
c^{(1,1)}_0 &  \cdots &  c^{(1,M)}_0 \\
\vdots & \ddots & \vdots \\
c^{(M,1)}_0 & \cdots& c^{(M,M)}_0
\end{array}
\right) =
\sum_{k = 1}^M \lambda_k v_k v_k^T,
\]
where $\lambda_k \in \reals$ and
$v_k \in \reals^M$ for $k = 1,\ldots, M$.
We choose as principal directions the eigenvectors $v_k$
associated with the $R$ largest eigenvalues.

\paragraph{Fit.}
We now describe how to fit the 
low-rank plus block diagonal kernel, starting
from an approximate kernel $\hat \Sigma$ fitted as explained in
\eqref{eq:gaussian_kernel_full}.
The kernel has the form
\BEQ
\label{eq:low_rank_block_diag_kernel}
\hat \Sigma^\text{lr+bd} = 
V^T \hat \Sigma^\text{lr} V + D,
\quad
D=
\left(
\begin{array}{ccc}
\bar C_1 & \cdots & 0 \\
  \vdots & \ddots & \vdots   \\
 0 & \cdots &   \bar C_M \\
\end{array}
\right),
\EEQ
where
$V \in \reals^{ R(P+F) \times M(P+F)}$,
$\hat \Sigma^\text{lr} \in \symm^{R(P+F)}$,
and
each $\bar C_i \in \symm^{(P+F)}$,
for $i = 1, \ldots, M$.
The sparse matrix $V$ is defined as 
\[
V = 
\left(
\begin{array}{ccccccc}
v_{1,1} & \cdots & 0 &\cdots & v_{1,M} & \cdots & 0 \\
\vdots &\ddots  & \vdots& \cdots&\vdots &\ddots  & \vdots  \\ 
0 &  \cdots & v_{1,1} &\cdots & 0 & \cdots  & v_{1,M}  \\ 
\vdots & \vdots & \vdots& &\vdots & \vdots & \vdots  \\ 
v_{R,1} & \cdots & 0 &\cdots & v_{R,M} & \cdots & 0 \\
\vdots &\ddots  & \vdots& \cdots&\vdots &\ddots  & \vdots  \\ 
0 &  \cdots & v_{R,1} &\cdots & 0 & \cdots  & v_{R,M}  \\ 
\end{array}
\right),
\]
that is, it projects the normalized residual,
concatenated in time according to 
\eqref{eq:multivariate_gauss},
along the $R$ principal directions,
concatenated in time in the same way.
Then, the low-rank approximation of the kernel
$\hat \Sigma^\text{lr}$ is given by
\[
\hat \Sigma^\text{lr} = 
V \hat \Sigma V^T.
\]
Finally, the block diagonal elements $\bar C_i$,
for $i = 1, \ldots, M$,
are chosen so that the block diagonal components
of $\hat \Sigma$
and
$\hat \Sigma^\text{lr+bd}$
are equal. That is,
\[
{(\bar C_m)}_{i,j} = 
{(\hat \Sigma - V^T \hat \Sigma^\text{lr} V)}_{i + (m-1)(P+F),
j + (m-1)(P+F)}
\]
for all
$i = 1, \ldots, (P+F)$,
$j = 1, \ldots, (P+F)$,
and 
$m = 1, \ldots, M$.
We note that both the $\hat \Sigma^\text{lr}$ matrix
and all the $\bar C_m$ matrices are Toeplitz.
Clearly, the low-rank plus block diagonal kernel so
constructed is symmetric, but is not guaranteed
to be positive definite. Again we rely on 
Tikhonov regularization during inference to
correct any non-positive definiteness.

\paragraph{Infer.}
To infer missing values of the normalized residual
we again solve equation \eqref{eq:schur},
where $\hat\Sigma$
is replaced by $\hat\Sigma^\text{lr+bd}$. 
We note here
an efficient way to obtain the inverse matrix
\[
{(\hat\Sigma^\text{lr+bd}_{\mathcal{O},\mathcal{O}} + \lambda I)}^{-1}.
\]
We apply the well-known Woodbury formula,
where
$
A = D_{\mathcal{O},\mathcal{O}}+  \lambda I
$
is the initial matrix
and
$
V^T_{\mathcal{O},\mathcal{O}} \hat\Sigma^\text{lr} V_{\mathcal{O},\mathcal{O}}
$
is the low-rank correction.
The inverse is given by
\[
{(V^T_{\mathcal{O},\mathcal{O}} \hat\Sigma^\text{lr} 
V_{\mathcal{O},\mathcal{O}} + A)}^{-1} = 
A^{-1} - 
A^{-1} 
V^T_{\mathcal{O},\mathcal{O}}
{\left(
{(\hat\Sigma^\text{lr})}^{-1} + 
V_{\mathcal{O},\mathcal{O}}
A^{-1}
V^T_{\mathcal{O},\mathcal{O}}
\right)}^{-1}
V_{\mathcal{O},\mathcal{O}}
A^{-1}.
\] 
We discuss below how this computational procedure
helps to reduce the cost of inference.

\paragraph{Memory and computational cost.}
Using the low-rank plus block diagonal approximate kernel
allows to reduce both the memory usage and the computational 
cost of inference from the model.
The memory required to store 
$\hat\Sigma^\text{lr+bd}$
is $O(MR)$ for $V$, 
$O(K^2(P+F))$ for $\hat \Sigma^{lr}$,
and $O(M(P+F))$ for $D$,
so it is linear in $M$, quadratic in $R$,
and linear in $P+F$.
The computational cost of fitting the low-rank plus block diagonal
does not change significantly from the
cost of fitting $\hat \Sigma$, which is still
the costlier part.
The cost of inference, again in the worst case,
is 
$O((P+F)^3M)$ to compute $A^{-1}$,
$O((P+F)^3R^3)$ to compute the other inverses,
and of lower order for all the matrix multiplications,
since also $\hat\Sigma^\text{lr+bd}_{\mathcal{U},\mathcal{O}}$
is low-rank plus sparse.
So, the computational cost of inference is
linear in $M$ and cubic in $R$ and $P+F$,
a great advantage with respect to the
original kernel.
Again, in practice, caching can  dramatically  
reduce the execution time.

\section{Greedy grid search}
\label{sec:ggs}
We now explain the simple algorithm we use to select values
of the hyper-parameters 
in the various components of our model.
The algorithm consists in
fitting the model on some train data
and measuring the model performance on some test data
for different choices of the hyper-parameters,
starting from the first ones in their provided ranges,
ordered by increasing complexity,
and trying iteratively the next ones.
The algorithm stops when 
it finds a local optimum
of the evaluated performance of the
model on the test data,
up to a user-defined \emph{search width}
of possible changes in the hyper-parameter choices. 

\paragraph{Fit and evaluate.}
Consider a model with
$H>0$ hyper-parameters, which we call
 $ h = (h_1, \ldots, h_H)$.
Each has 
a range of possible values,
$h_1 \in \mathcal{H}_1 = \{h_1^{(1)}, \ldots, h_1^{(V_1)} \}$,
\ldots,
$h_H \in \mathcal{H}_H= \{h_H^{(1)}, \ldots, h_H^{(V_H)} \}$,
so
$h \in \mathcal{H} =
 \mathcal{H}_1 \times \cdots\times \mathcal{H}_H $.
A fit procedure, for given train data, is a mapping 
\[
\mathcal{F}: \mathcal{H} \to \mathcal{M}
\]
to a certain, typically very large dimensional,
\emph{model space} $\mathcal{M}$.
The procedure to evaluate a model, for given test data, is a function
\[
\mathcal{E}:  \mathcal{M} \to\reals
\]
from the model space to the real numbers,
so that the lowest the value, the more accurate the model
on the given test data.
In this section we look at ways to approximately
minimize the function
\[
(\mathcal{E} \circ \mathcal{F}):  \mathcal{H} \to\reals,
\]
\ie, to find the combination of hyper-parameter
values so that the model fitted with that choice
 is the approximate minimizer of the model evalution 
 function.

\paragraph{Hyper-parameters.}
We rely on three basic features of the hyper-parameters.
First, for each one we need  
a discrete range of possible values.
Second, the spacing among such values must
 be such that choosing the next or previous value
changes the model complexity
 by approximately the same amount.
 (Here we do not define ``model complexity''
 formally. In information theory
    it would be, roughly speaking,
    the bit size of 
an efficiently compressed 
representation of the model.)
Last, the ranges are ordered by increasing
 complexity of the resulting model, 
so the first values give rise to the 
simpler models.
Typically, simpler models are
either
more regularized, so
 some regularization hyper-parameter has a
 higher value, or more parsimonious, \ie,
have fewer parameters to fit.


\paragraph{Iterative search.}
So, given a sequence
of $H>0$ hyper-parameters $h = (h_1, \ldots, h_H)$
and for each one a range of possible values,
$h_1 \in \{h_1^{(1)}, \ldots, h_1^{(V_1)} \}$,
\ldots,
$h_H \in \{h_H^{(1)}, \ldots, h_H^{(V_H)} \}$,
a search width $W \geq1$,
and a hyper-parameter evaluation function
$(\mathcal{E} \circ \mathcal{F}): \mathcal{H} \to \reals$,
the greedy grid search algorithm is as follows.
\BIT
\item We define a cursor vector $(c_1, \ldots, c_H)$
where $c_i \in \{1, \ldots, V_i\}$ for every $i = 1,\ldots, H$, 
and initially
$(c_1, \ldots, c_H) = (1, \ldots, 1)$.
\item We repeat until convergence:
\begin{enumerate}
\item For any combination 
of cursor values $c'$ such that $\|c - c' \|_1 \leq W$,
we compute
$\xi(c') = (\mathcal{E} \circ \mathcal{F})
(h_1^{(c_1')}, \ldots, 
h_H^{(c_H')})
$
For example, if $W = 1$, we vary each $c_i$
by summing or subtracting 1, 
unless we are at the boundary of a range, 
in which case we only subtract or add. 
If $W = 2$, we sum and subtract 2 to each variable,
and also sum and/or subtract 1 to any pair of variables, and so on.
\item Let $c^\star$ be the cursor value
 such that $\xi(c^\star)$ is smaller
 than all other $\xi(c')$ tried.
If it is not unique,
we pick the one such that $\|c^\star\|_1$
is smaller than all $\|c'\|_1$,
and if still not unique, 
we pick the one that is alphanumerically smaller.
It $c^\star = c$ we exit,
and return the hyper-parameters $h_1 = h_1^{c_1}$, \ldots, 
$h_H = h_H^{c_H}$.
Otherwise,
we set the new value of $c$ to the value of $c^\star$.
\end{enumerate}
\EIT
A simple caching mechanism allows the algorithm just
described to not
re-evaluate the same set of hyper-parameters twice.
The returned sequence of hyperparameters
gives the lowest model evaluation on the test set
 among all its neighbors, up to
an $\ell_1$ distance of $W$.

\paragraph{Re-fit.}
After having chosen the hyper-parameters $h^\star$
with the procedure just described,
with the provided data divided into
train and test sets,
we re-fit the model on the whole dataset.

\paragraph{Computational cost.}
The algorithm described has only one parameter,
the search width $W$, which can be provided by the user.
By default it has value $W=1$. The larger the search width,
the more combinations of hyper-parameters will be tried,
and eventually (for large enough $W$) the whole grid will 
be tested, so the algorithm becomes a full grid search.
This has a cost exponential in $H$.
If instead the user chooses a small value of $W$
the computational cost may be dramatically lower.
In fact, the cost of each iteration
of the algorithm is approximately $O(H^W)$.
The number of iterations is random, 
but is likely to not depend significantly on $H$,
and instead on the spacing of the hyper-parameter ranges.
So, we may think of this algorithm as a way to turn
a search exponential in $H$ 
into one that is linear (if $W = 1$), 
quadratic (if $W=2$), and so on.
The returned choice of hyper-parameters 
is not guaranteed to be the same that
would be returned by the full grid search,
but is likely to
err on the side of caution,
\ie, to be more regularized and/or parsimonious
(since we start the search from the most conservative point).

\paragraph{Example: seasonal baseline.}
As an example, the procedure just described can be used
(as is in our software package)
to choose the values of the hyper-parameters 
$K^\text{trend}$, 
$K^\text{day}$, 
$K^\text{week}$, 
$K^\text{year}$, 
whose ranges are given in \S\ref{sec:baseline}.
The initial values are $K^\text{trend}=K^\text{day} = K^\text{week} = K^\text{year} = 0$, corresponding to a constant baseline $b_t = \beta_0$.
Then we iteratively increase the number of harmonics for
each periodicity, daily, weekly, and annual, and 
either turn on or turn off the trend term.
 We return the combination
whose evaluated loss on the test set is lowest.
We note that this simple strategy can be superior
to human tuning in many practical cases.  
For example, one might not expect the power
production of a wind turbine to have any weekly seasonality,
and hence might be tempted to hard-code $K^\text{week} = 0$,
but maybe maintenance of the turbine is typically done 
on Sundays (because power is less expensive),
and so, statistically, the turbine produces less power on that day.
The greedy search would presumably notice that, set 
some $K^\text{week}>0$, and thus capture this effect in 
the baseline model, at a very low computational cost.

\section{Software library}
The software implementation of the algorithm is available
online at
\begin{center}
\url{https://github.com/enzobusseti/tsar}.
\end{center}
It is written in Python and depends on
the standard scientific libraries
\verb+numpy+, \verb+scipy+,
\verb+pandas+, and \verb+numba+, which is used
to compile certain operations in machine code, to 
speed them up.

\paragraph{Fit.}
The fit procedure of the model is performed 
by the model constructor function.
In the simplest case, the syntax is as follows
\begin{Verbatim}[frame=single]
from tsar import tsar
model = tsar(data=data, past=P, future=F)
\end{Verbatim}
where \verb+data+ is a dataframe indexed by a datetime column,
with $M$ columns of floating point numbers, or missing values,
and \verb+P+ and \verb+F+ are integers, for the 
$P$ and $F$ constants.
The model constructor divides the data into
train and test datasets, performs greedy grid searches for
all of the baseline models, and for the residual Gaussian process,
and then refits the model on the whole data with the 
hyper-parameter values obtained.
The resulting \verb+model+ object can be efficiently serialized.


\paragraph{Infer.}
Once a model has been fitted,
inference is performed as follows
\begin{Verbatim}[frame=single]
prediction = model.predict(data=new_data, prediction_time=t)
\end{Verbatim}
where \verb+new_data+ is a dataframe indexed by 
a datetime column and with the same columns as the dataframe
\verb+data+,
and \verb+t+ is a datetime variable 
 describing time $t$.
The returned \verb+prediction+ dataframe has a datetime index
with datetime values described by
 $t-P+1, \ldots, t+F$, 
 the same columns as the dataframe \verb+data+,
and no missing values:
All the values that were already present in \verb+new_data+
are copied over and the others are filled in with
the procedure of equation \eqref{eq:schur}, plus the
computed baseline value.

\paragraph{Hyper-parameters.}
For any hyper-parameter explained above
the user can provide a value to the constructor function, 
or none. In that case the hyper-parameter
is flagged for greedy grid search optimization, 
and finally set to the result of the search.
That is what happens with the syntax we showed above.
So the user can specify, or leave unexpressed,
any number of hyper-parameters. 
For example, the user can set $K^\text{trend} = 0$
for certain components, but let the greedy grid search
find the values of $K^\text{day}$, and so on.
If no hyper-parameter is left unexpressed 
no greedy grid search is performed,
 and the model is trained only once,
 without splitting the data into train and test sets.


\section*{Acknowledgments}
The author thanks 
Stephen Boyd, 
Emmanuel Candes, 
Trevor Hastie,
Michael Kochenderfer, 
Nicholas Moehle,
Robert Tibshirani,
and his other former colleagues,
for interesting discussions.

\bibliography{tsar}

\end{document}